\documentclass[journal,twoside]{IEEEtran}
\usepackage[noadjust]{cite}

\usepackage{color,soul}
\soulregister\cite7 
\soulregister\citep7 
\soulregister\citet7 
\soulregister\ref7 
\soulregister\pageref7 

\usepackage{soul}

\ifCLASSINFOpdf
  \usepackage[pdftex]{graphicx}
  \graphicspath{{./pdf/}{./jpeg/}}
  \DeclareGraphicsExtensions{.pdf,.jpeg,.png,.jpg}
\else
  \usepackage[dvips]{graphicx}
  \graphicspath{{./eps/}}
  \DeclareGraphicsExtensions{.eps}
\fi
\usepackage{subfigure} 
\usepackage{xcolor}
\usepackage{lineno}

\usepackage{amsmath,amssymb,amsthm}
\usepackage[hyphens]{url}
\usepackage{cases}
\usepackage{multirow}

\usepackage{pdfpages}

\begin{document}

%
\title{Tele-Operated Oropharyngeal Swab (TOOS) Robot Enabled by TSS Soft Hand for Safe and Effective COVID-19 OP Sampling}
%
%
%

\author{Wei Chen$^{\dag}$, 
        Jianshu Zhou$^{\dag}$$^{*}$, 
        Shing Shin Cheng,  
        Yiang Lu,
        Fangxun Zhong, 
        Yuan Gao, 
        Yaqing Wang,
        
        Lingbin Xue, 
        Michael C. F. Tong
        and Yun-Hui Liu
\thanks{This work was supported in part by National Key Research and Development Program of China under Grant 2018YFB1309300.
This work is supported in part by RGC via project 14202918, in part by the Natural Science Foundation of China under Grant U1613218, in part by the Hong Kong Centre for Logistics Robotics, and in part by the VC Fund 4930745 of the CUHK T Stone Robotics Institute.}
\thanks{$^{\dag}$Wei Chen and Jianshu Zhou are co-first authors in this work.}
\thanks{$^{*}$Corresponding author: Jianshu Zhou, jianshuzhou@cuhk.edu.hk.}
\thanks{Wei Chen, Jianshu Zhou, Shing Shin Cheng, Yiang Lu, Fangxun Zhong, Yuan Gao, Yaqing Wang and Yunhui Liu are with the T Stone Robotics Institute, the Department of Mechanical and Automation Engineering, The Chinese University of Hong Kong.}
\thanks{JS, Zhou and YH Liu are also with the Hong Kong Center for Logistics Robotics.
}%
\thanks{Lingbin, Xue and Michael C. F. Tong are with the Department of Otorhinolaryngology, Head and Neck Surgery, The Chinese University of Hong Kong.}
}

\maketitle


\begin{abstract}
The COVID-19 pandemic has imposed serious challenges in multiple perspectives of human life. 
To diagnose COVID-19, oropharyngeal swab (OP SWAB) sampling is generally applied for viral nucleic acid (VNA) specimen collection.
However, manual sampling exposes medical staff to a high risk of infection. 
Robotic sampling is promising to mitigate this risk to the minimum level, but traditional robot suffers from safety, cost, and control complexity issues for wide-scale deployment. 
In this work, we present soft robotic technology is promising to achieve robotic OP swab sampling with excellent swab manipulability in a confined oral space and works as dexterous as existing manual approach. 
This is enabled by a novel Tstone soft (TSS) hand, consisting of a soft wrist and a soft gripper, designed from human sampling observation and bio-inspiration. TSS hand is in a compact size, exerts larger workspace, and achieves comparable dexterity compared to human hand. 
The soft wrist is capable of agile omnidirectional bending with adjustable stiffness.
The terminal soft gripper is effective for disposable swab pinch and replacement.
The OP sampling force is easy to be maintained in a safe and comfortable range (throat sampling comfortable region) under a hybrid motion and stiffness virtual fixture-based controller. 
A dedicated 3 DOFs RCM platform is used for TSS hand global positioning.
Design, modeling, and control of the TSS hand are discussed in detail with dedicated experimental validations. 
A sampling test based on human tele-operation is processed on the oral cavity model with excellent success rate. 
The proposed TOOS robot demonstrates a highly promising solution for tele-operated, safe, cost-effective, and quick deployable COVID-19 OP swab sampling.
\end{abstract}

\begin{IEEEkeywords}
COVID-19 sampling, oropharyngeal swab robot, soft robot, teleoperation
\end{IEEEkeywords}

%

\section{Introduction}
%
%
%
%
\IEEEPARstart{T}{he} outbreak of COVID-19 pandemic leads to profound challenges in people's daily lives in different fields lastingly and globally \cite{yang2020combating}. 
Although one year has passed since its original breakout, there are still significant number of new cases and the global outlook on the pandemic remains pessimistic \cite{link2}. 
It is estimated that COVID-19 may continue for several years before the herd immunity can be achieved.

At the forefront of combating COVID-19, oropharyngeal viral nucleic acid (VNA) detection is critical to identify potential patients for their timely diagnosis, effective quarantine, and dedicated treatment \cite{mcintosh2020coronavirus}. 
However, due to the highly contagious nature of COVID-19, the manual face-to-face VNA sampling is a high-risk task to the medical doctors/nurses involved \cite{bandyopadhyay2020infection, sanche2020high}. 
To reduce infection risk, medical staffs have to be armed with heavy protection suits, goggles, and masks, which lead to severe burnout and psychological stress during long hours of work shift and yet do not eliminate the risk of infection. 
News of medical staffs’ COVID-19 infection and even death continue to be reported across the globe \cite{bandyopadhyay2020infection}. 
The risks of infection and mortality have raised major concern among medical staff and their family members \cite{bandyopadhyay2020infection, sanche2020high}.

To minimize the infection risk and reduce the job-related emotional and physical burden among medical staffs, robotic sampling is a promising solution to this challenge \cite{shen2020robots}. 
Robotic technology has been consistently implemented to replace humans during the execution of various dangerous tasks \cite{billard2019trends, kovach2017evaluation, yang2020keep}. 
There have been reports about robots being used to accelerate, assist, and automate the oropharyngeal (OP) and nasopharyngeal (NP) swab sampling process \cite{shen2020robots}. 
For example, a swab robot with sampling force control within 10-60 g has been developed for clinical application with preliminary efficacy validation \cite{li2020clinical}. 
A low-cost miniature robot to assist swab sampling has been reported with efficient and convenient deployability \cite{wang2020design}. 
A remote human-robot collaborative control strategy has been developed for swab sampling \cite{ying2020remote}. 
Attempts to mount specialized sampling tools on commercially available robot arms have also been reported to facilitate robotic sampling \cite{link13, link14}.

Although pioneering efforts have been made to apply robotic technology in OP/NP swab sampling, there are two major concerns about existing OP/NP swab robots. 
First, the safety concern towards the testee is evident when a bulky and rigid robot is manipulated inside the narrow oral cavity with delicate soft tissue. 
Second, the dexterity of the terminal end-effector is highly limited with most robots only employing a one-directional swab insertion and retraction. 
Compared to human operators, existing robots are significantly inferior in terms of safety, user comfort, manipulation dexterity, and potential sampling completeness.

To tackle the above challenges, soft robotics technology developed in recent years provides an avenue for safe and effective swab sampling enabled by its inherent compliance and simplified system complexity \cite{rus2015design, lee2017soft, cianchetti2018biomedical}. 
Besides, the dexterous motion of the human wrist can be aptly replaced by soft robotic wrist \cite{laschi2012soft, kurumaya2018modular}. 
An example has been reported by using a soft actuator to function as a novel swab \cite{xie2021tapered}. 
It provides an interesting perspective to oral sample collection enabled by soft robotic approach. However, this soft robotic swab is expected to require more careful scrutiny and approval before large-scale clinical adoption. 
Using soft robotic wrist and hand to replace the human operator in handling existing disposable swabs could be a more practical approach to promote soft robotic sampling in the near term.

In this work, we present soft robotic technology is promising to achieve robotic OP swab sampling with excellent swab manipulability in a confined oral space and works as dexterous as existing manual approach.
A teleoperated oropharyngeal swab (TOOS) robot is developed, which is constructed by two components, namely the TStone Soft (TSS) hand and a 
dedicated remote center of motion (RCM) platform. 
The sampling dexterity is enabled by the TSS hand, consisting of a soft wrist and a soft gripper, designed from human sampling observation and bio-inspiration. TSS hand is in a compact size, exerts larger workspace, and achieves comparable dexterity compared to human hand. 
The soft wrist is realized by a multi-backbone continuum soft robot with omnidirectional bending and linear elongation/contraction. 
The stiffness of the soft wrist is pre-adjustable to allow both safety insurance and forceful interaction based on the task requirement.
A two-fingered soft gripper is designed to pinch and release the disposable swab, thus facilitating swab replacement.
The TSS hand is mounted at the terminal of a 3-degree of freedom (DOF) RCM platform to enable the swab to 
approach the mouth cavity of the testee during global positioning. 
An intuitive tele-operation controller is developed to enable real-time teleoperation of the OP swab sampling process, guided by the visual feedback from the camera mounted at the tip of the TSS hand. 
A hybrid motion/stiffness virtual fixture is designed to ensure the sampling force within the comfortable sampling range.
Dedicated experiments have validated the efficacy of the tele-operation control, variable stiffness capability, safe interaction during sampling, human-assisted swab replacement, and complete sampling process in a human user demonstration. 
The system can be applied to versatile OP swab sampling scenarios with manual sampling comparable dexterity using disposable swabs, which provides a promising solution to reduce face-to-face sampling infection risk of the medical staff.

The highlights of the contribution are as follows:
\begin{enumerate}
  \item A TOOS robot is presented for safe and effective OP swab sampling using existing disposable swab teleoperated by medical staff.
  \item A novel TSS hand is proposed, which is one of the most dexterous robotic hands dedicated to OP swab sampling using disposable swabs. TSS hand achieves human hand comparable swab manipulation dexterity.
  The wrist stiffness can be pre-adjusted based on the task requirement and weight of terminal components.
  \item A teleoperation controller with a novel hybrid motion/force virtual fixture (H-VF) is designed with the consideration of real-time motion teleoperation based on visual feedback and and haptic constraint.
  Sampling safety and comfort can be ensured by the dedicated H-VF constraints.
\end{enumerate}

The remaining of this paper is organized as follows: 
Section II introduces the overall system design of TOOS robot, TSS hand, and tele-operation framework; 
Section III discusses the kinematics and stiffness modeling of the TSS hand; 
Section IV presents the complete tele-operation framework of the robot; 
Section V provides experimental validation of the tele-operation framework, TSS hand interaction force, swab sampling success rate, and human user demonstration; 
Section VI presents the discussion and conclusion.

\section{OVERALL TELE-OPERATED OROPHARYNGEAL SWAB (TOOS) ROBOT DESIGN}

\begin{figure}[!t] 
	\vspace{-0.0cm}   
	\setlength{\abovecaptionskip}{-0.0cm}    
	\setlength{\belowcaptionskip}{-10cm}    
    \centering
    \includegraphics[width=\linewidth]{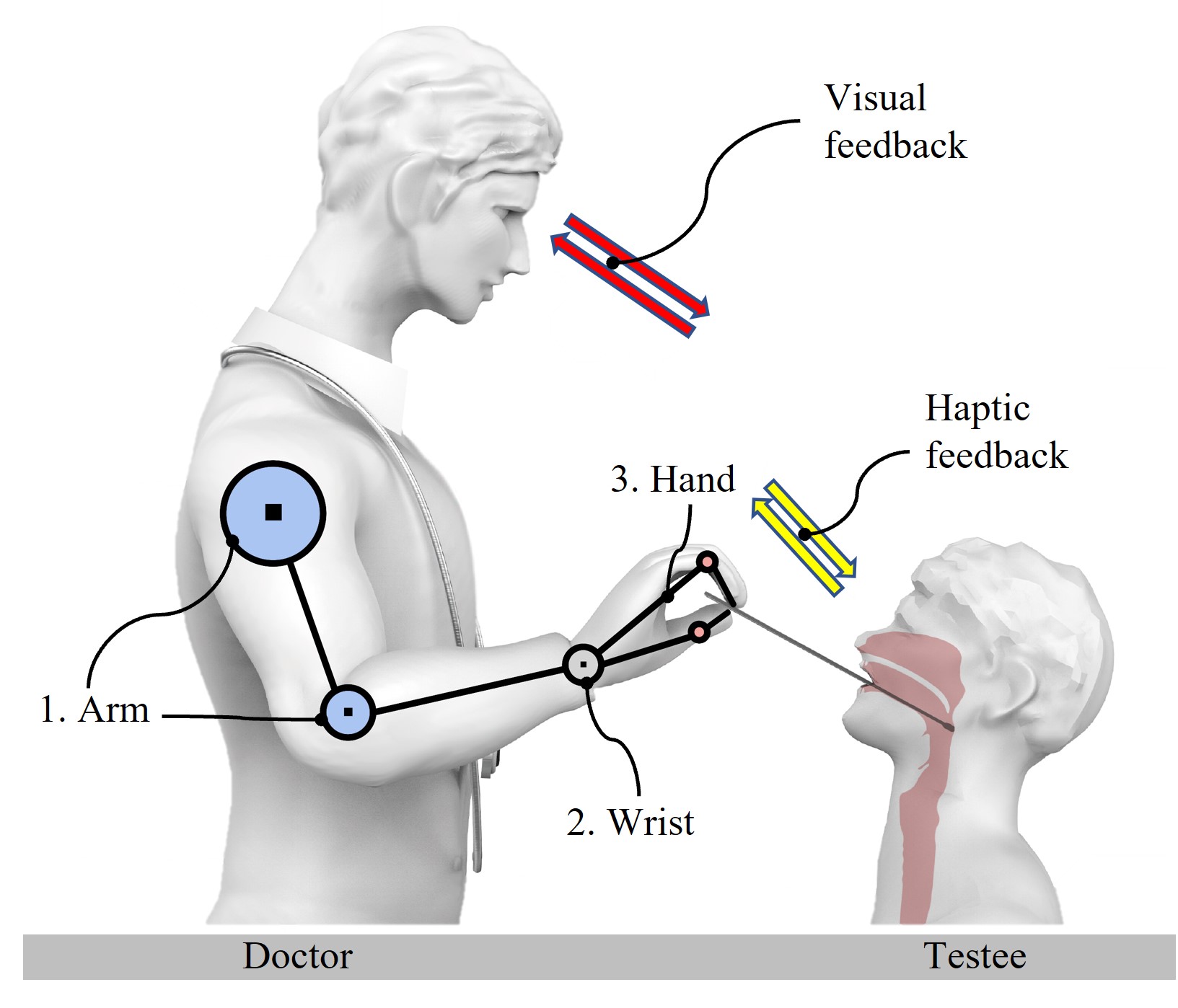}
    \caption{
        Healthcare staffs manual oropharyngeal swab sampling illustration. 
        Three physical interaction constructions (arm, wrist, and fingers) and two sensory perception constructions are included (visual and haptic feedback).
    }
    \label{fig:f1}
	\vspace{-0.2cm}   
\end{figure}

This section introduces the overall system design of the proposed tele-operated sampling system. 
The design intention is to replicate the physical joints and sensory perception of a human operator that are involved in manual sampling process.

\begin{figure*}[!th] 
	\vspace{-0.0cm}   
	\setlength{\abovecaptionskip}{-0.0cm}    
	\setlength{\belowcaptionskip}{-10cm}    
    \centering
    \includegraphics[width=\linewidth]{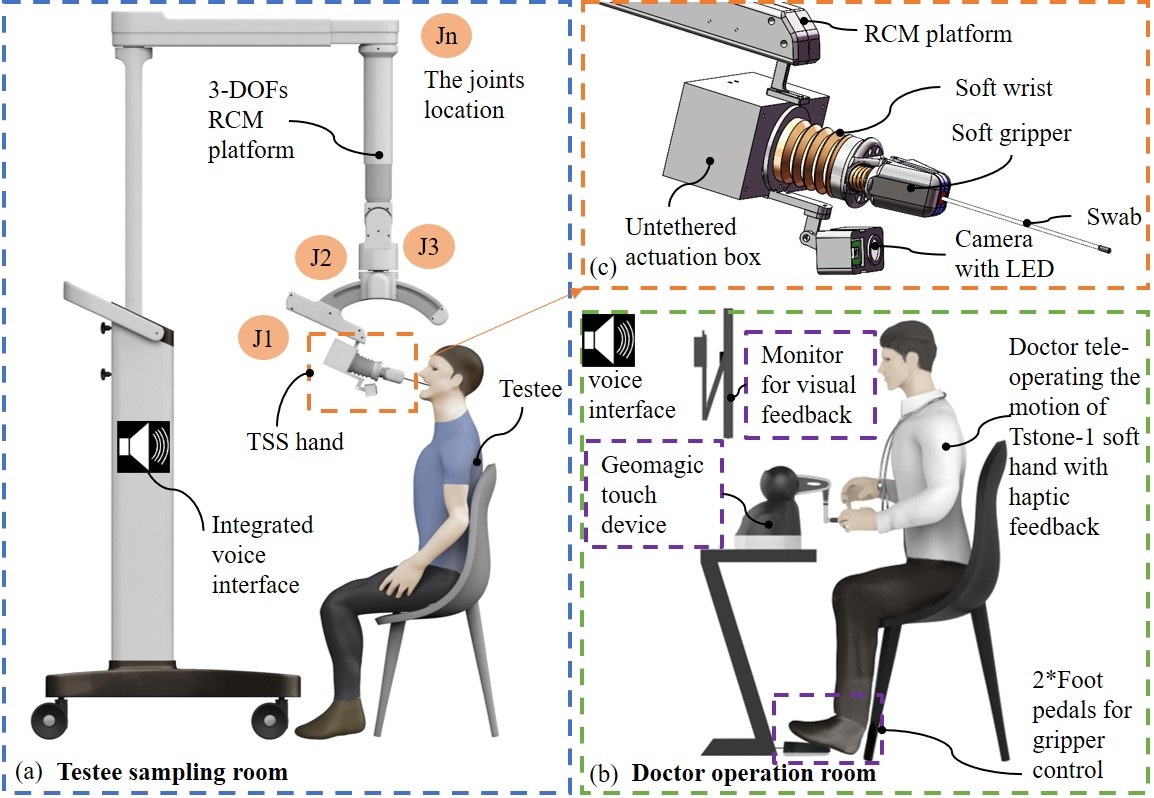}
    \caption{
        Overview of the proposed tele-operated oropharyngeal swab (TOOS) sampling system. 
        (a) The test sampling room with the TOOS robot and testee. 
        The TOOS robot is composed of a 3-DOF medical platform and a Tstone soft (TSS) hand. 
        (b) Doctor operation room with doctor and teleoperation systems based on terminal camera feedback. 
        (c) TSS hand majorly composed of a soft gripper and a soft wrist.
    }
    \label{fig:f2}
	\vspace{-0.2cm}   
\end{figure*}

\subsection{Manual Sampling Observation and Its Inspiration}

To derive our robot design, we investigated how well-trained healthcare workers perform the OP swab sampling.
By observing manual sampling process, there are physically three main functional components \cite{link21, petruzzi2020covid}, as shown in Fig. \ref{fig:f1}.
The first functional component is the arm that brings the hand close to the subject's mouth. 
The second function is realized by adjusting the swab direction with the wrist. 
The third one is to use fingers (especially the thumb, index finger, and middle finger) to tightly pinch and manipulate the swab. 
Considering the sensing units, the manual sampling process involves two key sensing components.
One is the human eye that can generate visual information, and another is the tactile sense of the human hand providing haptic feedback \cite{link21, petruzzi2020covid}.
Thus, our objective is to achieve safe and efficient robotic-assisted sampling by realizing these two sensory capabilities and three physical functions with dedicated robotic technology.
Furthermore, using the robot to implement the actuation and sensing functions as similar as possible to the human approach will provide the benefits of operation intuitiveness and convenience.

\subsection{TOOS Robot Design}

Following the human sampling study, the three physical maneuvers are intended to be achieved by two robotic parts. 
A 3-DOF RCM platform functions as the human arm to globally position the terminal components approaching the testee \cite{zhong2019foot} while a dedicated bio-inspired TSS hand enables the dexterous wrist motion and swab picking functions similar to the human hand \cite{zhou2017soft, zhou2018bcl, zhou2019soft}.
As illustrated in Fig. \ref{fig:f2}, the OP swab sampling is performed in two separate rooms/locations for testee sampling (Fig. \ref{fig:f2} (a)) and doctor tele-operation (Fig. \ref{fig:f2} (b)), respectively.
The testee sits in front of the TOOS robot and prepares to open the mouth for sampling while the doctor 
operates the master device based on the visual and haptic feedback to remotely manipulate the TSS hand. The doctor and the testee can communicate through the remote voice interaction.

The platform provides a global positioning of the terminal TSS hand.
The three DOFs distribution is illustrated in Fig. \ref{fig:f2} (a) as $J_n$, where $n$ is the number and location of the related DOF. 
$J_1$ is the insertion prismatic joint with 100 mm of motion range;
$J_2$ is the sagittal plane rotation joint with 47 degrees of motion range; 
$J_3$ is the frontal plane rotation joint with 360 degrees of motion range. 
The initial configuration of the platform can be arranged to adapt to the posture of a human sitting on a chair with the mouth angled up by 30 to 45 degrees. 
More specific configuration adjustment can be achieved for individual testee by remotely controlling each joint of the robot.

After the global positioning, the RCM platform is fixed with the TSS hand robot. 
After the orientation about the RCM is confirmed, $J_2$ to $J_3$ are locked. 
$J_1$ then provides the insertion and extraction motion of the TSS hand, which is controlled by virtual buttons in the dedicated HMI, which will be further discussed in Section. IV. 
After the swab is approaching the desired sampling target observed by medical staff, the hybrid motion/stiffness virtual fixture will be enabled by pressing the virtual fixture button. 
Then, the dexterous sampling motion is achieved by the TSS hand, which motion is tele-operated by the doctor via the Geomagic haptic device and the open/close is controlled by the foot pedals.

\begin{figure}[!tb] 
	\vspace{-0.0cm}   
	\setlength{\abovecaptionskip}{-0.0cm}    
	\setlength{\belowcaptionskip}{-10cm}    
    \centering
    \includegraphics[width=0.9\linewidth]{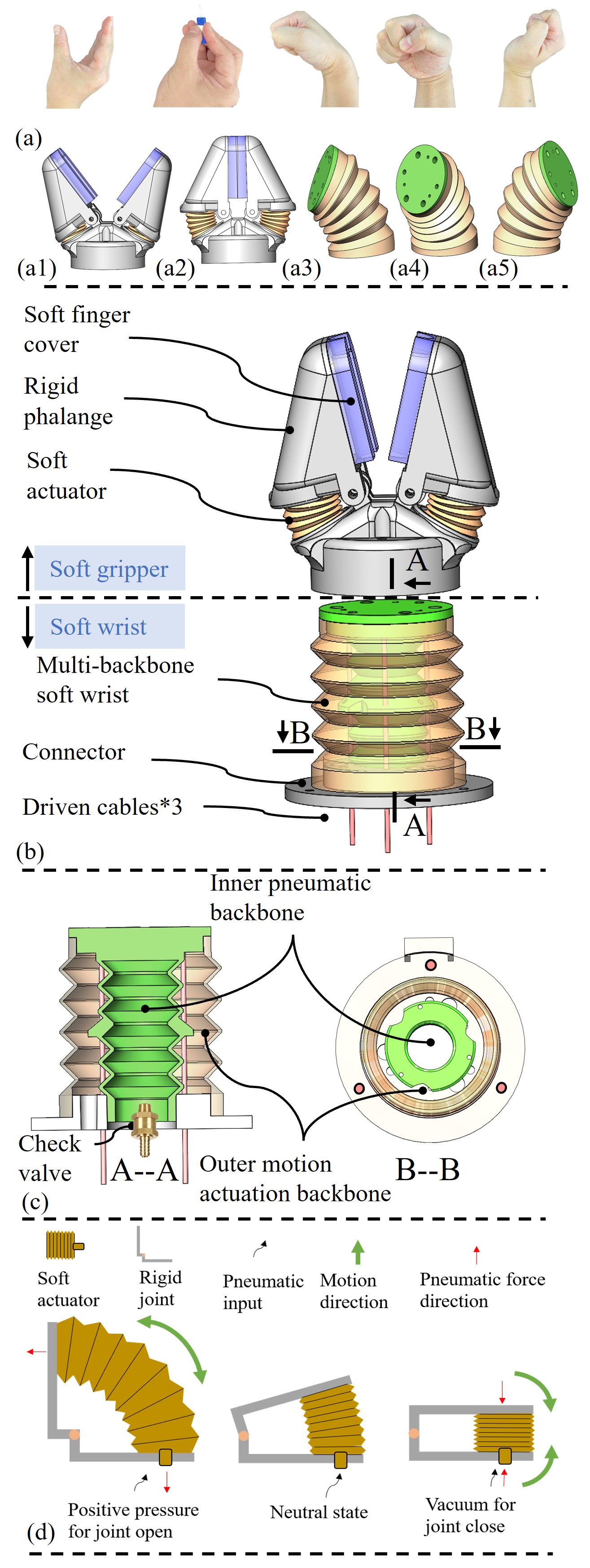}
    \caption{
        TSS hand design.
        (a) Intended human hand comparable swab manipulation dexterity for TSS hand. 
        (a1-a2) Swab pinch and release.
        (a3-a5) Dexterous wrist motion.
        (b) TSS hand construction.
        (c) Soft wrist design and construction.
        (d) Hybrid gripper working mechanism. 
    }
    \label{fig:f3}
	\vspace{-0.5cm}   
\end{figure}

\begin{table}[!bh]
\begin{center}
\caption{MAJOR PARAMETERS OF THE PROTOTYPE TSS SOFT HAND}
\begin{tabular}{|p{3cm}|p{3cm}|}
    \hline   
    \multicolumn{1}{|c}{\centering $Feature$} &
    \multicolumn{1}{|c|}{\centering $Parameter$} \\
    \hline   
    Soft gripper material & 3D printed PLA \\
    \hline
    Soft wrist material & 3D printed TPU \\
    \hline
    Height of the gripper & 80 mm \\
    \hline
    Length of the wrist & 65 mm \\
    \hline
    Diameter of the wrist & 60 mm \\
    \hline
    Weight of the hand & 160 g \\
    \hline
    Weight of the control box & 330 g \\
    \hline
    Waterproof & Yes \\
    \hline
\end{tabular}
\end{center}
	\vspace{-0.2cm}   
\end{table}

\subsection{TSS Hand Design}

\begin{figure}[!tb] 
	\vspace{-0.0cm}   
	\setlength{\abovecaptionskip}{-0.0cm}    
	\setlength{\belowcaptionskip}{-10cm}    
    \centering
    \includegraphics[width=\linewidth]{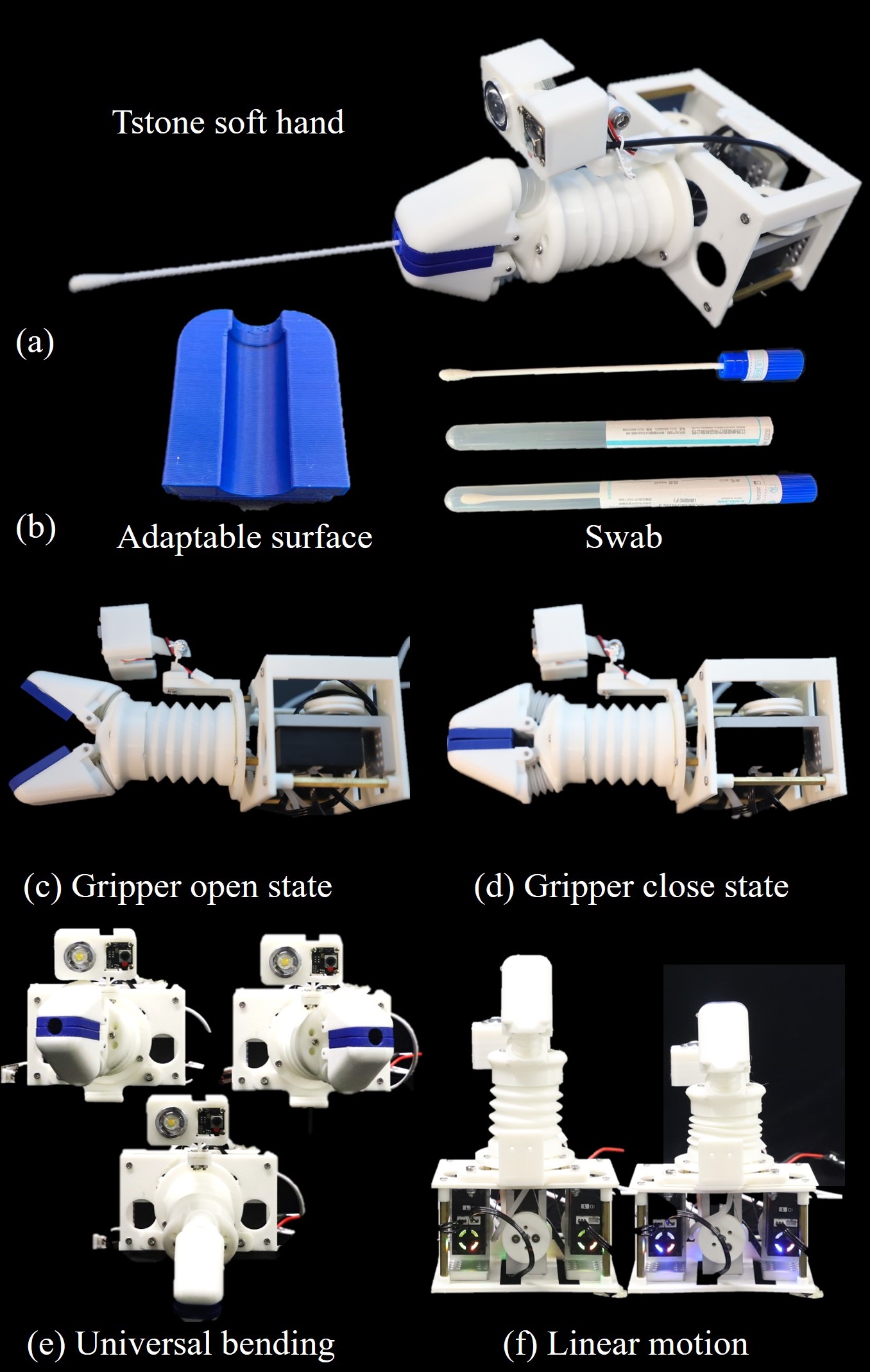}
    \caption{
    Real photos of the prototype TSS hand. 
    (a) Tstone-1 hand picking a swab. 
    (b) The soft finger cover adapts to the shape of swab tip. 
    (c) Original state of the Tstone hand. 
    (d) Gripper open state of TSS hand. 
    (e, f) Universal bending of TSS hand. 
    (g) Linear motion performance of TSS hand.}
    \label{fig:f4}
	\vspace{-0.2cm}   
\end{figure}

The major challenge is to achieve the human hand comparable swab manipulation dexterity, as shown in Fig. 3(a), in a compact, lightweight, and safe robotic hand. A novel TSS hand is designed to tackle this challenge \cite{zhou2017soft, zhou2018bcl, zhou2019soft, zhou2018soft, zhou2020adaptive}.
The TSS hand is designed in a human hand similar configuration and connected at the terminal of the RCM platform, as seen in Fig. \ref{fig:f3}. 
It consists of a universal soft wrist and a soft gripper.
The soft gripper is designed by a hybrid soft-rigid structure with soft joints and rigid phalanges to achieve both grasping compliance and robustness \cite{zhou2017soft, zhou2018bcl, zhou2019soft}. 
The soft joint is a pneumatic origami actuator with internal lumen and can elongate and shrink upon application of positive and negative pneumatic actuation as depicted in Fig. 4(d).  
The cover of the finger contact area is made of deformable material and has a cylindrical slot to fit the shape of the swab holder as depicted in Fig. \ref{fig:f4} (b). 
The gripper is designed to have an initial closed state. 
Pneumatic pressure is then applied to produce the actuation force. 
The gripper opens and closes when the negative pressure and positive pressure are applied, respectively, as shown in Figs. \ref{fig:f4} (c) and (d). 
The grasping force can be adjusted and controlled by the magnitude of the applied positive pressure \cite{zhou2019grasping, zhou2018intuitive}. 
The hybrid soft gripper provides merits of lightweight, easy control, high compliance and robustness for swab grasping and releasing.

As for the soft wrist design, it is intended to achieve multi-directional bending as dexterous as a human wrist in a confined space. Besides, the stiffness of the wrist is better to be adjustable to ensure both sampling safety and forceful interaction.
The multi-backbone continuum soft robot is a good candidate to serve this intention. 
The body of the wrist is composed of two nested backbones in an origami structure, with one backbone functions as stiffness tuning backbone and one backbone serves as motion actuation backbone \cite{zhou2020proprioceptive, mcmahan2005design}. 
Actuated by three driving cables, the outer motion actuation backbone provides dexterous motion of the robot.  
When the three cables are actuated simultaneously, the robot can extend and contract its body to adjust the length of the wrist.
The stiffness of soft wrist can be pre-adjusted towards different tasks. 
The stiffness of soft wrist determines the interaction force indirectly and influences the kinetic stability of the TSS hand. 
Therefore, the variable stiffness capability is helpful to provide suitable interaction force and desired kinetic stability for a specific task. 
This capability is enabled by the pneumatic adjustment of its inner stiffness tunning backbone. 
The stiffness tuning backbone functions as a pneumatic spring, the stiffness of which can be adjusted by the applied pressure. 
The variable stiffness characteristic provides the safety and payload guarantee at the same time depending on the task requirements.

The actual prototype pictures of the TSS hand are shown in Fig \ref{fig:f4}. 
Fig. \ref{fig:f4} (a) shows the TSS hand firmly holding a swab. 
The adaptable cover of the gripper and the commercially available swab are depicted in Fig. \ref{fig:f4} (b). 
Fig. \ref{fig:f4} (c) shows the open state of the TSS hand while Fig \ref{fig:f4} (d) depicts the initial closed state of the hand. 
Fig. \ref{fig:f4} (e) shows the universal bending of the wrist and Fig. \ref{fig:f4} (f) presents the linear motion of the wrist.

Enabled by the soft wrist and soft gripper, TSS hand is one of the most dexterous robotic hands for OP swab sampling with human hand comparable dexterity. Major design parameters of the TSS hand are provided in Table I.
The construction of the whole hand is 3D printable. 
The gripper was printed with PLA (RAISE3D N2 PLA 1.75 mm) while the softcover of the gripper and the wrist were printed with TPU (CR-TPU 1.75 mm). 
Three commercially available servo motors (Hiwonder LX-15D) and one portable pneumatic source \cite{zhou2019soft} are used to actuate the robot.
Furthermore, the hand is fully waterproof and therefore convenient for disinfection between sampling on different testees.

\subsection{Visual and Haptic Feedback}

We integrated a camera to provide the human eye's visual feedback and a haptic device to provide the human hand's haptic feedback. 
The camera is a commercially available 5 mega-pixel product, which works with an LED, as shown in Fig. 2(c) and Fig. 4(a), for brightness compensation of the testee's throat. 
Besides, a Geomagic touch haptic device, as shown in Fig. 2(b), is used to provide the haptic feedback and also as the master device to tele-operate the TSS hand.

\section{TSS HAND MODELLING}


To achieve effective teleoperation of the TSS hand, the hand model needs to be derived. For the soft gripper, we adopt the on/off control for swab grasping and releasing. 
For the universal soft wrist, we have to derive the kinematics model from the actuation space to the task space for forward/inverse kinematics-based tele-operation.
Besides, the stiffness of the universal soft wrist can be calibrated as a function of the pneumatic input, allowing pre-adjustment of its stiffness in the controller.

\begin{figure}[!t] 
	\vspace{-0.0cm}   
	\setlength{\abovecaptionskip}{-0.0cm}    
	\setlength{\belowcaptionskip}{-10cm}    
    \centering
    \includegraphics[width=\linewidth]{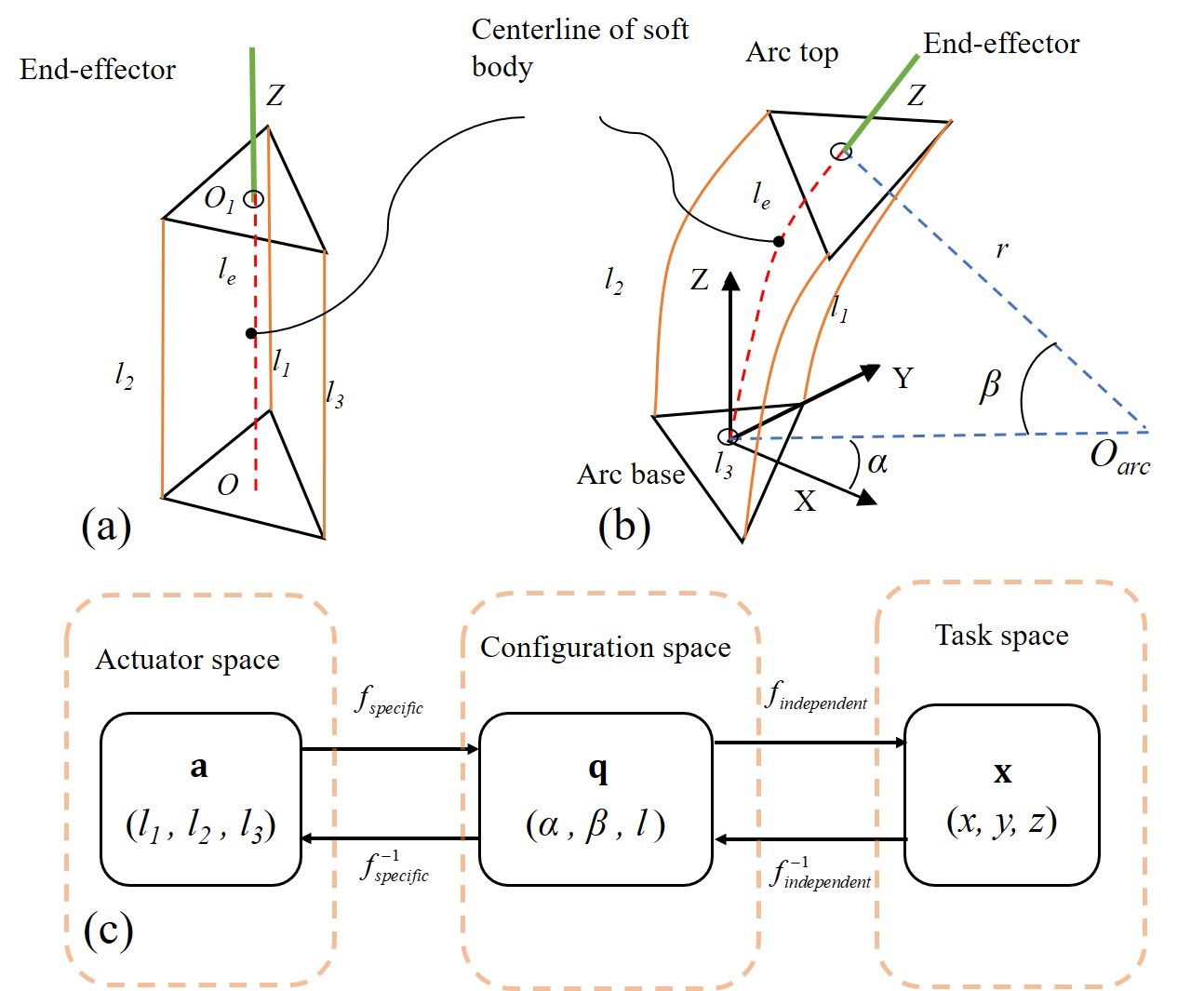}
    \caption{
        Kinematics modeling of the proposed soft wrist. 
        (a) The soft wrist at linear state. 
        (b) The soft wrist at bending state. 
        (c) Kinematics transformation among actuator, configuration, and task space.
    }
    \label{fig:f5}
	\vspace{-0.2cm}   
\end{figure}

\subsection{Soft Wrist Kinematics Modeling}

Firstly, we derive the robot-specific kinematics which maps the length of the driving cables to the major parameters of the robot configuration space. The position of the robot is quantified by the point on the centerline, as illustrated by the red dashed line in Figs. \ref{fig:f5} (a) and (b). As the body of soft wrist is continuum, we can express the position of the centerline by the length of driven cables \cite{huang2015sensor, camarillo2008mechanics, webster2010design}.

The length of centerline $l$ is the average length of the three driven cables. 
The curvature and place can be calculated by the trigonometric relationship  \cite{zhou2019grasping, zhou2018intuitive}. 
Then, the configuration space parameter $\mathbf{q} = (\alpha, \beta, l) \in \{\mathbf{q}\}$ can be expressed by the actuator space parameter $\mathbf{a} = (l_1, l_2, l_3) \in \{\mathbf{a}\}$ as:

\begin{equation}
    \setlength{\abovedisplayskip}{4pt}
    \setlength{\belowdisplayskip}{4pt}
    \begin{aligned} 
        \alpha &= \tan^{-1}(\frac{l_1 + l_3 - 2l_2}{\sqrt{3}(l_1 - l_3)}) \\
        \beta &= \frac{\sqrt{l_1^2 + l_2^2 + l_3 ^ 2 - l_1l_2 - l_1l_3-l_2l_3}}{d(l_1 + l_2 + l_3)} \\
        l &= \frac{l_1 + l_2 + l_3}{3} \\
    \end{aligned} 
\end{equation}

Considering that the final position of the gripper is of a distance Z away from the top center $O_1$.
Then, we derive the robot independent kinematics, which maps the configuration space parameters to the task space parameters, the position of gripper.

\begin{multline}
    \setlength{\abovedisplayskip}{4pt}
    \setlength{\belowdisplayskip}{4pt}
    \mathbf{T}_{end-effector} = \\ 
    \begin{bmatrix} 
        c_{\alpha}^2 c_{\beta} + s^2_{\alpha} & c_{\alpha}c_{\beta}s_{\alpha} - c_{\alpha}s_{\alpha} & c_{\beta} & Z c_{\alpha} s_{\beta} + \frac{l}{\beta}(1-c_{\beta})c_{\alpha} \\ 
        c_{\alpha}c_{\beta} - c_{\alpha} & s^2_{\alpha}c_{\beta} & s_{\beta} & Z s_{\alpha} s_{\beta} + \frac{l}{\beta}(1-c_{\beta})s_{\alpha} \\ 
        -c_{\beta} & -s_{\alpha} & c_{\beta} & Z  c_{\beta} + \frac{l}{\beta}s_{\beta} \\
        0 & 0 & 0 & 1  
    \end{bmatrix}
\end{multline}

After the forward kinematics derivation, we derive the inverse kinematics of the TSS hand. 
The basic concept is using the position of gripper to derive the length of the three driven cables. 
To derive the inverse kinematics, we first derive the configuration parameters \cite{dupont2009design, jones2006kinematics, gong2021soft}, as follows:

\begin{equation}
    \setlength{\abovedisplayskip}{4pt}
    \setlength{\belowdisplayskip}{4pt}
    \begin{aligned} 
        \alpha &= \tan^{-1}(\frac{y}{x})  \\
        \beta &=
        \begin{cases}
            2 \tan^{-1}(\frac{y}{z \sin{\alpha}}) & \text{if } \alpha \neq k \pi \\
            2 \tan^{-1}(\frac{x}{z \cos{\alpha}}) & \text{otherwise }
        \end{cases} \\
        l &=
        \begin{cases}
            2 \tan^{-1}(\frac{x}{z \cos{\alpha}}) &  \text{if } \alpha \neq k \pi \\
            0 & \text{otherwise }
        \end{cases}
    \end{aligned} 
\end{equation}

The length of each individual driving cable is then calculated as follows:
\begin{equation}
    \setlength{\abovedisplayskip}{4pt}
    \setlength{\belowdisplayskip}{4pt}
    l_i = l - \beta \cos(\frac{2\pi}{3}(i-1) + \frac{\pi}{2} - \alpha), i = \{1, 2, 3\}
\end{equation}

\subsection{Universal Wrist and Swab Terminal Stiffness Modelling}

The stiffness of the universal soft wrist can be adjusted by the pneumatic input of the inner pneumatic backbone. 
The larger the input pressure, the larger the stiffness of the wrist.
In this work, the relationship between the axial stiffness ($k_{wrist-axial}$) and lateral stiffness ($k_{wrist-lateral}$) of the wrist is experimentally calibrated as a function of the input pressure P. 
In-depth analytical modelling of the stiffness variation mechanism will be developed in our future work.

The interaction force during sampling is also highly dependent on the stiffness of the applied swab, $k_{swab-axial}$ and $k_{swab-lateral}$, which can be experimentally determined. 
Finally, the effective sampling stiffness, $k_{axial}$ and $k_{lateral}$, can be expressed as a function of the wrist stiffness and the swab stiffness under stiffness integration based on the parallel spring mechanism \cite{sackfield2013mechanics}. 
The effective stiffness $k_{axial}$ and $k_{lateral}$ can then be expressed as:
\begin{equation}
    \setlength{\abovedisplayskip}{4pt}
    \setlength{\belowdisplayskip}{4pt}
    \begin{bmatrix}
        k_{axial} \\
        k_{lateral}
    \end{bmatrix}
    =
    \begin{bmatrix}
        \frac{k_{wrist-axial}k_{swab-axial}}{k_{wrist-axial} + k_{swab-axial}} \\
        \frac{k_{wrist-lateral}k_{swab-lateral}}{k_{wrist-lateral} + k_{swab-lateral}}
    \end{bmatrix}
\end{equation}

\section{TELE-OPERATION FRAMEWORK BASED ON HYBRID motion/force VIRTUAL FIXTURE (H-VF)}

With the kinematic control and stiffness modulation, a tele-operation framework has been developed for OP sampling. 
This section discusses the teleoperation framework from system construction, controller design, to operation interface. 
The overall tele-operation framework can be categorized into three subsections, namely the RCM platform tele-operation, the TSS hand tele-operation based on H-VF, and the dedicated human-machine interface (HMI) for the tele-operation control.

\subsection{RCM Platform Tele-operation}

\begin{figure*}[!t] 
	\vspace{-0.0cm}   
	\setlength{\abovecaptionskip}{-0.0cm}    
	\setlength{\belowcaptionskip}{-10cm}    
    \centering
    \includegraphics[width=\linewidth]{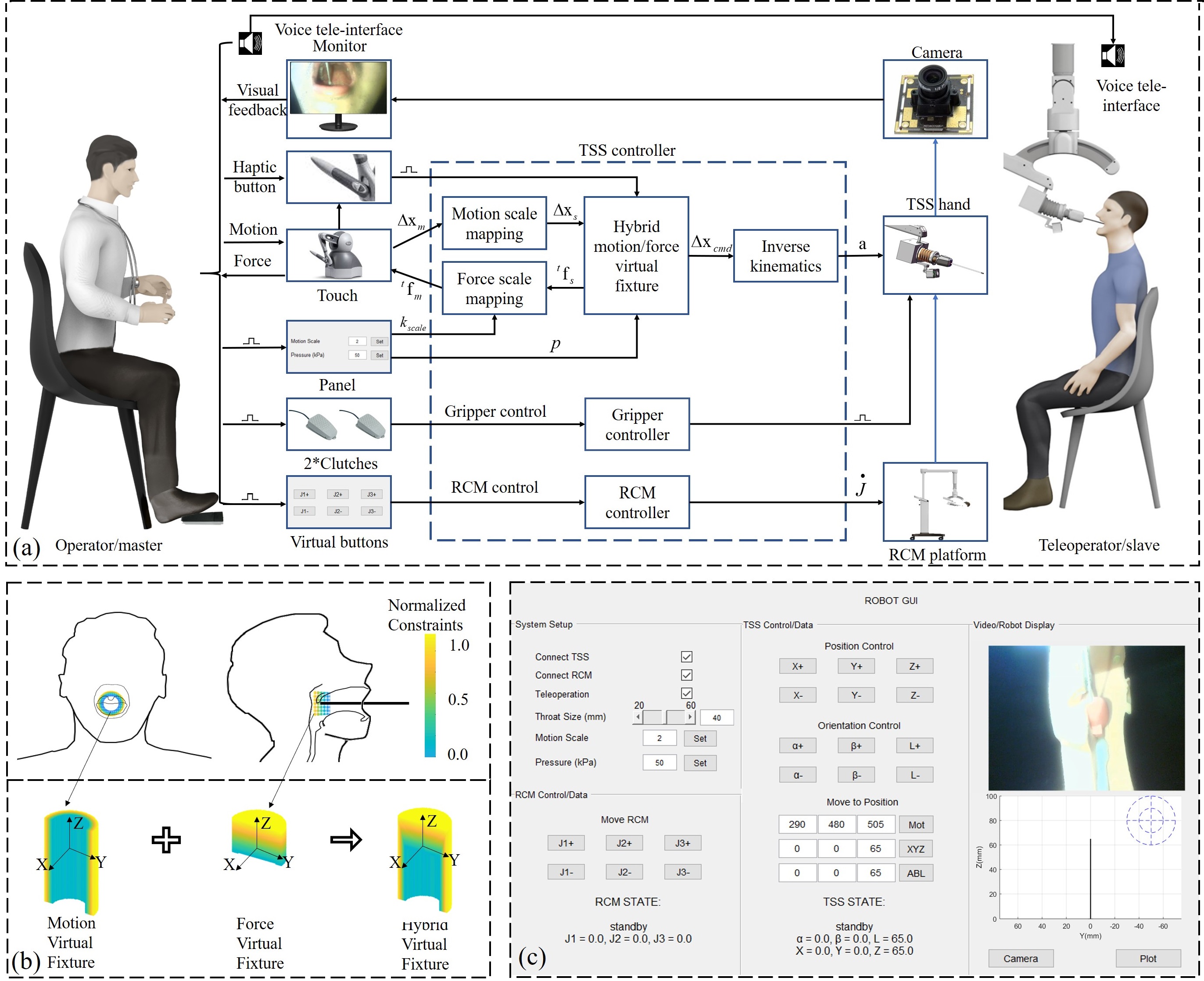}
    \caption{
        Overview of the applied teleoperation framework based on visual and haptic feedback. 
        (a) The overall teleoperation controller block diagram, which is majorly constructed by TSS controller, RCM controller, and gripper controller. 
        (b) Hybrid virtual fixture (H-VF) illustration. 
        (c) A dedicated human-machine interface (HMI) consisting of system setup, stiffness pre-adjustment, motion scale adjustment, RCM control panel, TSS control panel, visual feedback, and robot state illustration.
    }
    \label{fig:f6}
	\vspace{-0.2cm}   
\end{figure*}

As shown in Fig. \ref{fig:f6} (a), the medical staff operates the master system on the left side. 
The RCM platform is initially adjusted based on the sitting position of the testee. 
The configuration of RCM platform can be further adjusted by a point-to-point mapping controller (Fig. \ref{fig:f6} (a)) using the virtual buttons (Fig. \ref{fig:f6} (c)), which control the motion of the three DOFs of the platform. Then the swab can be inserted into the mouth of the testee and retracted after sampling \cite{song2006tele, mohammadi2016cooperative}.

The RCM platform controller is embedded in an onboard computer and the motor drivers are connected via CAN-bus communication. 
The controller is designed to run on the Windows operating system with a dedicated graphical user interface as shown in Fig. \ref{fig:f6} (c). 
The control signal is transferred by USB serial port.

\subsection{TSS Hand Tele-operation based on H-VF Control}
In terms of the TSS hand teleoperation, a TSS controller is designed to enable point-to-point motion control from master to slave system. 
A motion scale factor $k_{scale}$ is applied to adjust the motion ratio between the master and slave system. 
A hybrid motion/force virtual fixture (H-VF) is applied in the TSS controller to ensure sampling safety.

The logic of the TSS controller is illustrated in the middle section of the Fig. \ref{fig:f6} (a).
The master motion, $\Delta \mathbf{x}_m \in \mathbb{R}^3$, is obtained from the human operated haptic device. 
The mapping from the master system to the slave system is described as:

\begin{equation}
    \setlength{\abovedisplayskip}{4pt}
    \setlength{\belowdisplayskip}{4pt}
    \Delta \mathbf{x}_s = \frac{1}{k_{scale}} {}^s \mathbf{J}_{m} \Delta \mathbf{x}_m
\end{equation}
where the robot motion target $\Delta \mathbf{x}_s \in \mathbb{R}^3$ is converted by the motion scale factor $k_{scale}$.
This motion scale adjustment helps to restrict the movement of the swab in a confined throat space.
When the motion scale factor $k_{scale}$ is set to be $2$, the converted motion target $\Delta \mathbf{x}_s$ is half of the operator's motion $\Delta \mathbf{x}_m$. 
${}^s \mathbf{J}_{m}$ is the motion coordinate transformation matrix and pre-calibrated as:

\begin{equation}
    \setlength{\abovedisplayskip}{4pt}
    \setlength{\belowdisplayskip}{4pt}
    {}^s \mathbf{J}_{m} = 
    \begin{bmatrix}
        0 & 1 & 0 \\
        1 & 0 & 0 \\
        0 & 0 & -1 
    \end{bmatrix}
\end{equation}

After that, the robot motion command is processed by the hybrid virtual fixture (H-VF) controller.

The hybrid motion/force virtual fixture is illustrated in Fig. \ref{fig:f6} (b), which is programmable constraints and can provide double safety insurance for the OP sampling process \cite{park2001virtual, abbott2003virtual, li2007telerobotic, yang2019adaptive}.
The virtual fixture can be divided into two constraints: motion constraint and stiffness constraint.
In task space, since the swab sampling force is always in the comfortable range (60 g force), the motion virtual fixture is applied to restrict any large unsafe movement of the swab inside the human mouth (in the X-Y plane).
The boundary of the motion virtual fixture is determined based on the visual detection of the face width of the testee, as shown in Fig. \ref{fig:f6} (b). 
The operator is trained to estimate the physical dimensions of the oral cavity based on how it is observed from the camera when the swab touches throat surface. This observation-based dimension estimation then determines the 2D (X-Y) boundary of the motion virtual fixture. When the swab is seen to be in the center of the patient's throat from the camera view, the motion virtual fixture is triggered manually to set the origin of the virtual fixture. This approach provides an approximate registration between the testee's and the robot's coordinate system. The compliance of soft hand and swab can compensate for the registration bias in this version. The registration performance can be enhanced by involving computer vision assistance through leveraging binocular camera and RGBD image processing in our future work. 
In the configuration space, when the interaction force will surpass the comfortable range (60 g force) after certain deformation \cite{li2020clinical}, the stiffness virtual fixture is applied as shown in Fig. \ref{fig:f6} (b).
Our hybrid motion/force virtual fixture can be defined as an optimisation problem \cite{li2007telerobotic}:

\begin{equation}
    \setlength{\abovedisplayskip}{4pt}
    \setlength{\belowdisplayskip}{4pt}
    \begin{aligned} 
        & \Delta \mathbf{x}_{cmd} = \arg \min_{\Delta x} \Vert \Delta \mathbf{x}_s - \Delta \mathbf{x} \Vert_2 \\
        & s.t. \qquad x_s^2 + y_s^2 \leq r_{throat}^2 \\
        & \quad \quad l_s \leq l_{button} + l_{stiffness}
    \end{aligned} 
\end{equation}
where $\Vert \Delta \mathbf{x}_s - \Delta \mathbf{x} \Vert_2$ is the objective function when $\Delta \mathbf{x}_s$ is applied to the robot.
$x_s^2 + y_s^2 \leq r_{throat}^2$ ($\Delta \mathbf{x} = (x_s, y_s, z_s)^T$) represents the motion constraint conditions in task space and $r_{throat}$ is the throat size which is selected by the doctors/nurses.
$l_s \leq l_{button} + l_{stiffness}$ represents the stiffness constraint conditions in configuration space, which depends on the axial stiffness as $l_{stiffness} = f_{safety} / k_{axial}$ (here $f_{safety}$ is 60 g) and length of the robot when the button is pressed ($l_{button}$).
$l_s$ is length of the robot mapping $\Delta \mathbf{x}$ in task space.
The output of H-VF is the robot motion command in task space ($\Delta \mathbf{x}_{cmd} \in \{\mathbf{x} \}$), which is converted to the configuration space command ($\mathbf{q}_{cmd} \in \{\mathbf{q} \}$) and then to the actuation space command ($\mathbf{a} \in \{ \mathbf{a}\}$) by using the inverse kinematics and to be delivered to the driven motors, as illustrated in Section III-A and Fig. {\ref{fig:f6}} (a).
The intersection area between the motion and stiffness boundaries is the hybrid motion/stiffness virtual fixture for safe sampling, as shown in Fig. \ref{fig:f6} (b). 
The virtual fixture can be timely enabled or disabled by pressing the switch button on the Geomagic device.
When the medical staff observing the swab terminal point approaches or contacts the desired throat surface, the H-VF can be triggered (which means $l_{button}$ is selected) by the operator as shown in Fig. \ref{fig:f6} (a).
A dedicated tele-operation UI is designed as illustrated in Fig 6(c). The robot state and camera feedback can be observed in this UI. Furthermore, this UI provides motion VF range adjustment from 20 to 60 mm diameter (double of $r_{throat}$) for the operator.

In terms of hardware realization, the TSS robot controller is built on a laptop with MATLAB on the Windows operating system. 
The cable is driven by three servo motors, controlled through the USB serial port.

\subsection{Haptic Constraint Force Calculation and Realization}

In terms of haptic constraint for operator, the constraint force is used to constrain the operator's motion to ensure the motion of swab within the region of hybrid virtual fixture. 
The haptic constraint force is calculated by the consideration of both motion constraint and stiffness constraint of robot.
The motion virtual fixture provides the constraint force ${}^t \mathbf{f}_{motion} \in \mathbb{R}^3$ in task space, which is calculated by the impedance model:

\begin{equation}
    \setlength{\abovedisplayskip}{4pt}
    \setlength{\belowdisplayskip}{4pt}
    {}^t \mathbf{f}_{motion} = k_{motion} \Delta \mathbf{\zeta}
\end{equation}
where $k_{motion}$ is the predefined motion feedback stiffness (here $0.5 N/mm$ is selected for a smooth haptic device operation based on subjective experience). 
$\Delta \mathbf{\zeta} = \Delta \mathbf{x}_{cmd} - \Delta \mathbf{x}_s \in \mathbb{R}^3$ is the motion difference when robot motion target $\Delta \mathbf{x}_s$ exceeds the virtual motion boundary.

The stiffness virtual fixture constraint force ${}^q \mathbf{f}_{stiffness}$ $\in$ $\mathbb{R}^3$ in configuration space is calculated by the stiffness model of the TSS hand:

\begin{equation}
    \setlength{\abovedisplayskip}{4pt}
    \setlength{\belowdisplayskip}{4pt}
    {}^q \mathbf{f}_{stiffness} = k_{stiffness} \Delta \mathbf{\xi}
\end{equation}
where $k_{stiffness}$ is a pre-defined stiffness for safety and smooth operation based on subject experience (here $0.5 N/mm$ is selected).
$\Delta \mathbf{\xi} = \mathbf{q}_{cmd} - \mathbf{q}(\Delta \mathbf{x}_s) \in \mathbb{R}^3$ is the deformation of the robot when the virtual surface is touched.
$\mathbf{q}(\Delta \mathbf{x}_s)$ is the configuration space state mapping $\Delta \mathbf{x}_s$ in task space. 
This force is not the force that applied on the robot tip, but the force calculated from the stiffness model intended for safety constraint.

The overall constraint force, ${}^t \mathbf{f}_s \in \mathbb{R}^3$, in the robot task space can be calculated as:

\begin{equation}
    \setlength{\abovedisplayskip}{4pt}
    \setlength{\belowdisplayskip}{4pt}
    {}^t \mathbf{f}_s = {}^t \mathbf{f}_{motion} + {}^t \mathbf{J}_{q} {}^q \mathbf{f}_{stiffness}
\end{equation}
where ${}^t \mathbf{J}_{q}$ is the jacobian mapping the force from the configuration space $\{\mathbf{q}\}$ to the task space $\{\mathbf{x}\}$.

Finally, the force is transferred to the master system to function as the operation haptic constraint force as:

\begin{equation}
    \setlength{\abovedisplayskip}{4pt}
    \setlength{\belowdisplayskip}{4pt}
    {}^t \mathbf{f}_m = {}^m \mathbf{J}_{s} {}^t \mathbf{f}_s
\end{equation}
where ${}^m \mathbf{J}_{s} = {}^s \mathbf{J}^T_{m}$ is the constraint force coordinate transformation matrix.

For the haptic constraint realization, the haptic device communicates with TSS controller through UDP protocol.

\subsection{Soft Wrist Stiffness Pre-adjustment}
The stiffness of soft wrist can be pre-adjusted towards different tasks. 
There are two major influential factors to the stiffness value. 
One is the desired interaction force. 
The smaller the interaction force is required, the smaller the original stiffness of the wrist should be. The second factor is the open-loop motion control accuracy and kinetic stability. 
The larger the stiffness of the wrist, the higher the teleoperation accuracy and kinetic stability, because the higher stiffness will reduce the passive deformation result from the weight of terminal components.
Usually, the stiffness value is a tradeoff between these two factors for the desired task.

In this sampling task, the stiffness of our used disposable swab is sufficiently compliant to ensure the sampling safety, which will be evaluated in the experimental section. 
Thus, the stiffness of the wrist is majorly determined by the motion accuracy consideration. 
A relatively high stiffness value will provide a more accurate tele-operation. 
The specific stiffness value can be adjusted in the upper left corner of the HMI as shown in Fig. \ref{fig:f6} (c), where the portable pneumatic actuation box provides -30 to 90 kPa pressure adjustment range.

\subsection{Dedicated Human-Machine Interface (HMI) for Teleoperation}
To tele-operate the robot effectively, the above-discussed control is operationally realized by a dedicated HMI, as shown in Fig. \ref{fig:f6} (c).
The dedicated HMI is composed of the actuation space panel, configuration space panel, video streaming feedback, robot real-time state, and stiffness 
adjustment panel. 
The configuration space panel is connected to the haptic device for human hand motion capture to control the robot. 
The HMI is illustrated in real-time at the monitor for visual feedback of the operator.

\section{EXPERIMENTAL VALIDATION} 

\begin{figure}[!t] 
	\vspace{-0.0cm}   
	\setlength{\abovecaptionskip}{-0.0cm}    
	\setlength{\belowcaptionskip}{-10cm}    
    \centering
    \includegraphics[width=\linewidth]{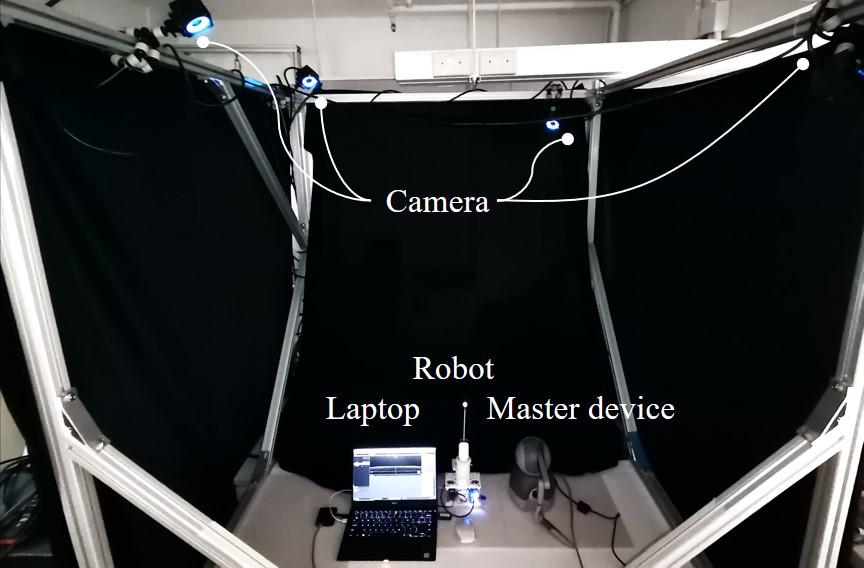}
    \caption{
        The optical tracking system setup for teleoperation accuracy test.
    }
    \label{fig:f7}
	\vspace{-0.2cm}   
\end{figure}

This section provides the comprehensive experimental validation of the whole proposed TOOS sampling system, 
including the teleoperation efficacy validation, swab terminal sampling force validation, sampling success rate validation, and human user demonstration.

\subsection{Tele-Operation and Hybrid Motion/Stiffness Virtual Fixture Validation}
The experiment setup to validate the tele-operation efficacy is depicted in Fig. \ref{fig:f7}, where the master device (Haptic operator) and the slave device (TSS hand) were placed on a table. 
Four OptiTrack cameras were hung at four corners of the experimental setup, enclosed by black curtains to ensure the motion capture accuracy. 
By using the motion capture software (Motive, OptiTrack), the optical trackers
were calibrated with mean error of 0.196 mm. 
The pressure of robot was set at 90 kPa for this experiment to enhance the motion accuracy and kinetic stability.

TSS hand achieves a human hand comparable workspace and is capable of dexterous omnidirectional bending within the workspace.
As shown in Fig. 8(a), the yellow region represents the TSS hand and the blue region represents the human hand (Pronation: 75.8 mm, Supination: 82.1 mm, Flexion: 76.4 mm, Extension: 74.9 mm, Radial deviation: 21.5 mm, Ulnar deviation: 36.0 mm \cite{boone1979normal}). 
Half of their workspaces are depicted for comparison convenience, and the whole workspaces are almost symmetrical.
As a result, TSS hand can realize dexterous swab manipulation in a human hand comparable workspace, which benefits the teleoperation intuitiveness and ensures the swabbing performance.

\begin{figure}[!tb] 
	\vspace{-0.0cm}   
	\setlength{\abovecaptionskip}{-0.0cm}    
	\setlength{\belowcaptionskip}{-10cm}    
    \centering
    \includegraphics[width=\linewidth]{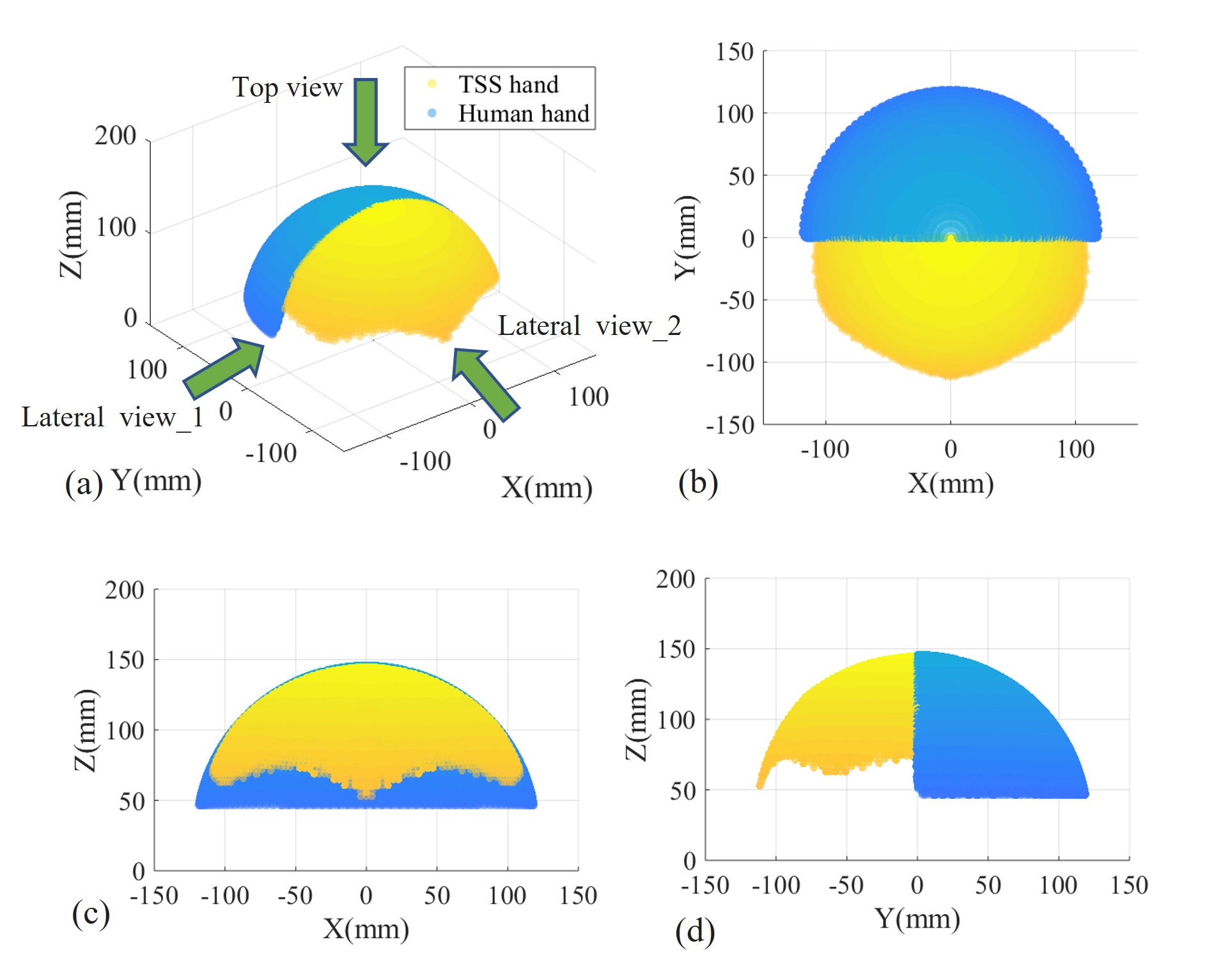}
    \caption{
        TSS hand and human hand workspace comparison. 
        (a) 3D view. 
        (b) Top view. 
        (c) Lateral view-1.
        (d) Lateral view-2.
    }
    \label{fig:f8}
	\vspace{-0.2cm}   
\end{figure}

\begin{figure}[!tb] 
	\vspace{-0.0cm}   
	\setlength{\abovecaptionskip}{-0.0cm}    
	\setlength{\belowcaptionskip}{-10cm}    
    \centering
    \includegraphics[width=\linewidth]{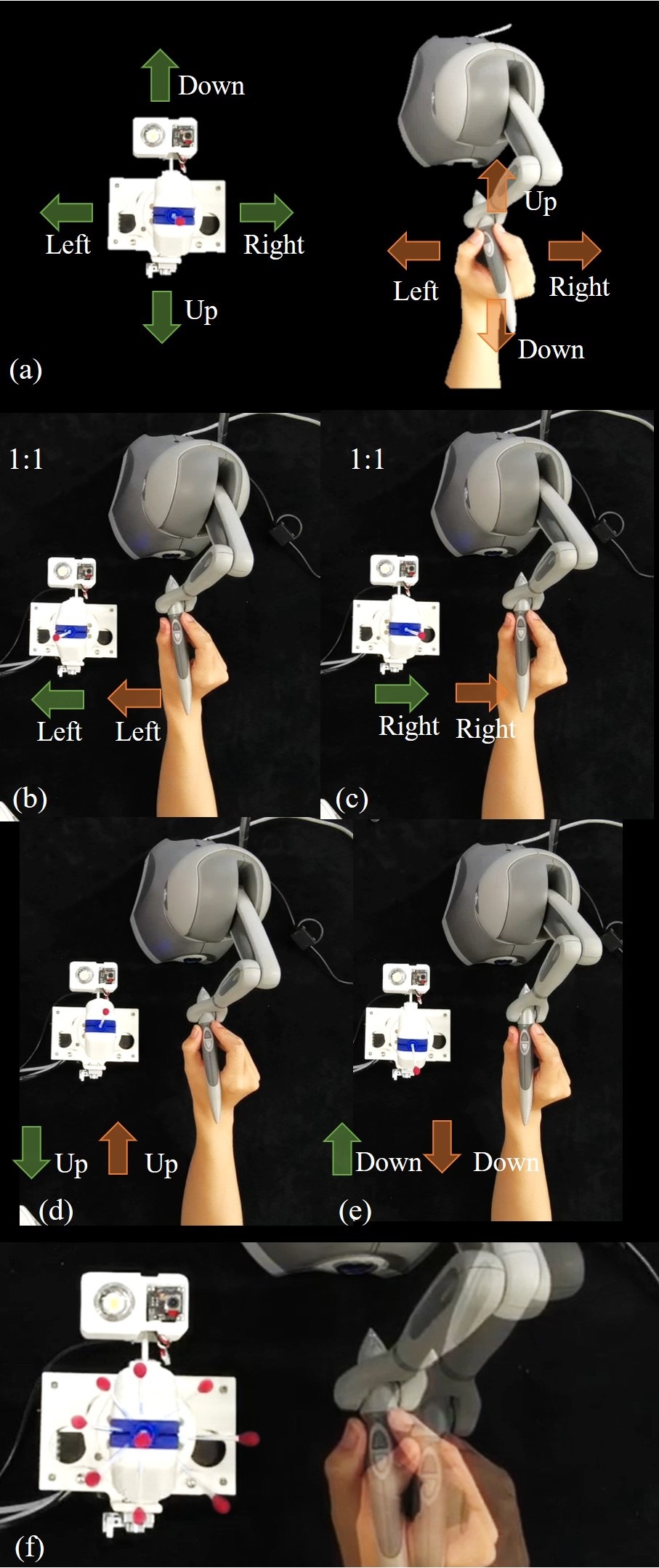}
    \caption{
        Teleoperation efficacy test 
        (a) Direction illustration of master device and robot. 
        (b-e) Left, right, up, and down directional tele-operation. 
        (f) Revolving trajectory tele-operation.
    }
    \label{fig:f9}
	\vspace{-0.2cm}   
\end{figure}

\begin{figure*}[!tb] 
	\vspace{-0.0cm}   
	\setlength{\abovecaptionskip}{-0.0cm}    
	\setlength{\belowcaptionskip}{-10cm}    
    \centering
    \includegraphics[width=0.95\linewidth]{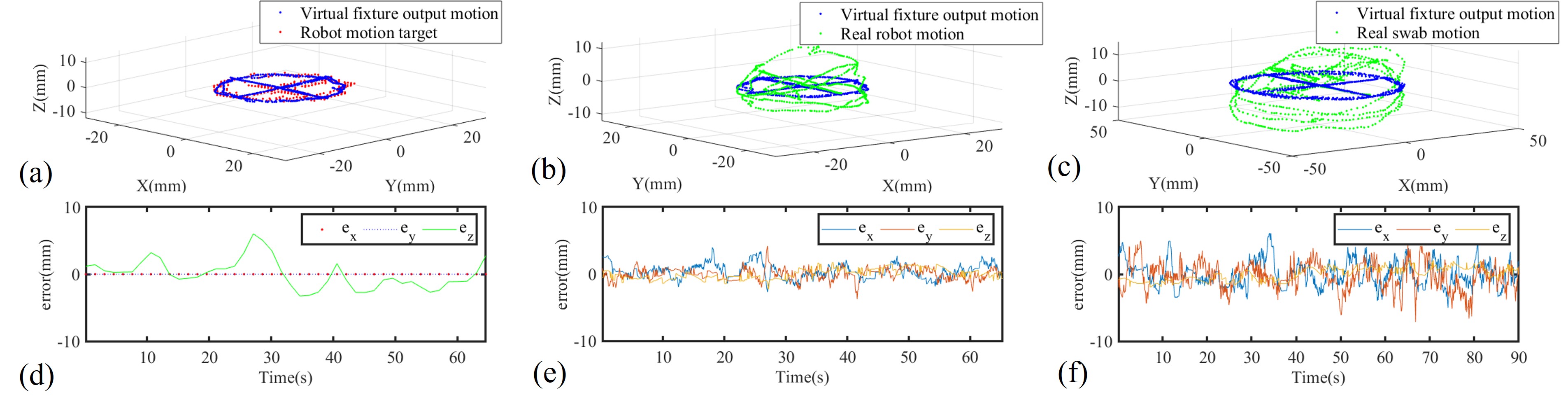}
    \caption{
        Tele-operation accuracy and hybrid vitural fixture efficacy validation. 
        (a) Trajectory point clouds of robot motion target and output motion of virtual fixture. 
        (b) Trajectory point cloud of virtual fixture output motion and real robot motion. 
        (c) Trajectory point cloud of virtual fixture output motion and real swab motion. 
        (d) Error of (a) in three dimensions. 
        (e) Error of (b) in three dimensions. 
        (f) Error of (c) in three dimensions.
    }
    \label{fig:f10}
	\vspace{-0.2cm}   
\end{figure*}

The tele-operation performance is presented in Fig. \ref{fig:f9}. 
As the robot is front and back reversely posed, the real motion direction, in the sampling process, of both master and slave are labeled as green and yellow arrows, respectively, in the figure. 
Figures \ref{fig:f9} (b)-(e) present the tele-operation efficacy of the front, back, up, and down movements of the TSS hand. 
The TSS hand was able to follow the human operator movement in real-time. Besides, a random continuum trajectory teleoperation was processed with the results shown in Fig. \ref{fig:f9} (f). 
The left side of the figure shows the feature points of the TSS hand trajectory throughout the process. 
The TSS hand was able to smoothly follow the motion of the human hand within 0.04 s delay under 25 Hz control frequency.

To evaluate the accuracy performance of the tele-operation, we recorded the motion of the human hand by the haptic device, the trajectory output of the virtual fixture controller, and the captured real motion of the TSS hand in 
the teleoperation process. 
The accuracy of the tele-operation can be analyzed by the comparison of these three trajectories. 

As shown in Fig. \ref{fig:f10} (a), the robot motion target $\Delta \mathbf{x}_s$ is depicted as the red dotted line, and the virtual fixture output $\Delta \mathbf{x}_{cmd}$ trajectory is depicted as the blue dotted line. 
A comparison is made between them in 3D space, and the position error in each dimension is depicted in Fig. \ref{fig:f10} (d). 
In X-Y plane, since the human hand motion was within the motion virtual fixture, there was no position error. 
In the Z direction, the robot was restricted by the force virtual fixture, leading to the position difference observed in Fig. \ref{fig:f10} (d).

Then, we compared the virtual fixture output trajectory with the TSS hand tip trajectory. 
The trajectory point cloud comparison is depicted in Fig. \ref{fig:f10} (b). 
The position errors in each direction of X, Y, and Z are depicted in Fig. \ref{fig:f10} (e).
The root mean square error in all of the three axes is (1.1864, 0.8660, 0.7579) mm, which shows an acceptable teleoperation accuracy.

A similar comparison was made between the virtual fixture output trajectory and the swab tip real trajectory. 
The point cloud of these two trajectories is depicted in Fig. \ref{fig:f10} (c). 
Position errors in three dimensions are illustrated in Fig. \ref{fig:f10} (f). 
The root mean square position error in three dimensions is (1.9996, 2.0037, 0.9708) mm.
The increment of error is caused by the increased length of the swab compared to the tip of the TSS hand, which amplifies the position error of the TSS hand.
All the dynamic performance of teleoperation performance was recorded and presented in the accompanying video attachment.

\begin{figure}[!bt] 
	\vspace{-0.0cm}   
	\setlength{\abovecaptionskip}{-0.0cm}    
	\setlength{\belowcaptionskip}{-10cm}    
    \centering
    \includegraphics[width=\linewidth]{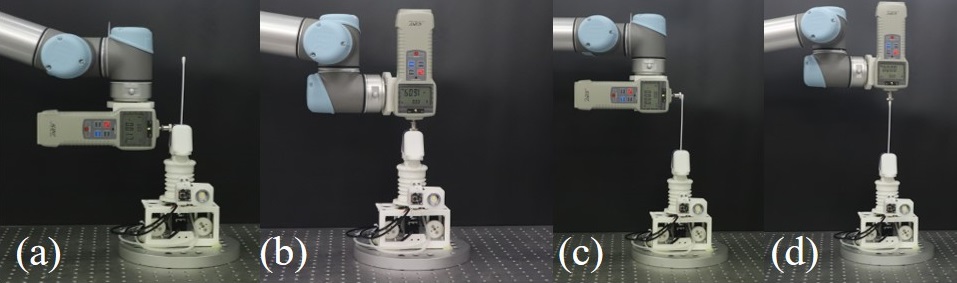}
    \caption{
        Force experiment. 
        (a) TSS hand lateral stiffness test. 
        (b) TSS hand axial stiffness test. 
        (c) Swab lateral interaction force test. 
        (d) Swab axial interaction force test.
    }
    \label{fig:f11}
	\vspace{-0.2cm}   
\end{figure}

\begin{figure}[!bt] 
	\vspace{-0.0cm}   
	\setlength{\abovecaptionskip}{-0.0cm}    
	\setlength{\belowcaptionskip}{-10cm}    
    \centering
    \includegraphics[width=\linewidth]{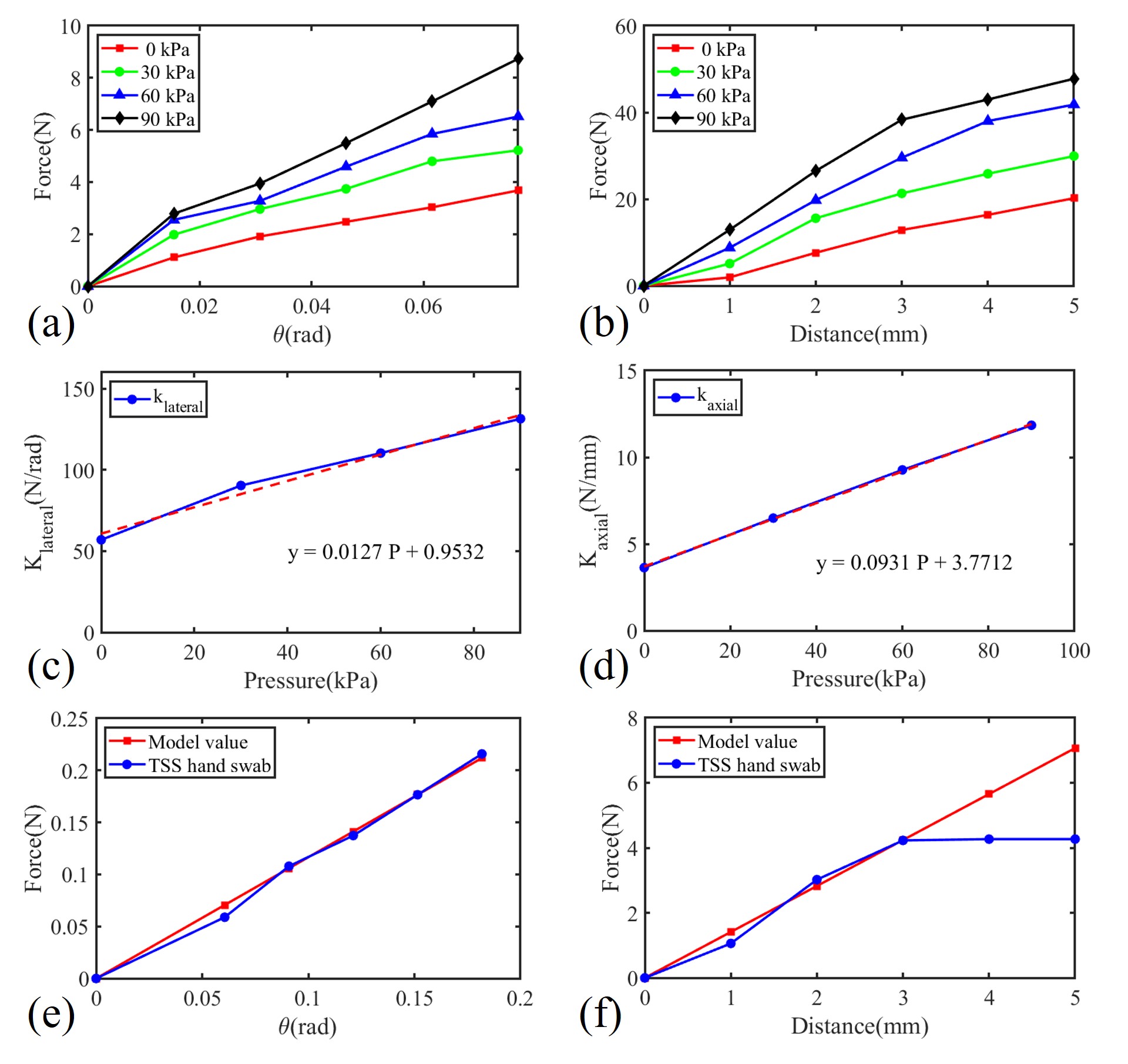}
    \caption{
        Sampling force test results 
        (a) TSS hand lateral stiffness test result. 
        (b) TSS hand axial stiffness test result.
        (c) TSS hand lateral stiffness fitting and fitting function. 
        (d) TSS hand axial stiffness fitting and fitting function. 
        (e) Swab lateral interaction force test result (90 kPa soft wrist). 
        (f) Swab axial interaction force test result (90 kPa soft wrist).
    }
	\label{fig:f12}
	\vspace{-0.2cm}   
\end{figure}

The effective sampling forces in both axial and lateral directions were measured under different input pressures with the swab pinched by the TSS hand, as shown in Figs. \ref{fig:f11} (c) and \ref{fig:f11} (d). 
As the swab we used was quite flexible, the stiffness of soft wrist has limited influence on the terminal stiffness of the swab. 
The lateral interaction force of the swab was always kept in a compliant, safe, and comfortable level even under the maximum wrist stiffness under 90 kPa pressure as shown in Fig. \ref{fig:f12} (e). 
The maximum interaction force was 0.002 N (20 g) which is within the reported comfortable sampling force of 60 g \cite{li2020clinical}. 
The interaction force in the axial direction will exceed the comfortable range. For example, the axial interaction force reached 4.23 N (423 g) under 3 mm passive deformation of the swab as shown in Fig \ref{fig:f12} (f). 
However, this condition can be avoided by our (motion) virtual fixture controller.

\begin{table}[!bth]
\begin{center}
\caption{Swab terminal stiffness under different TSS hand stiffness}
\begin{tabular}{|p{2.2cm}|p{1.4cm}|p{0.7cm}|p{0.7cm}|p{0.7cm}|p{0.7cm}|}
    \hline   
    \multicolumn{2}{|c}{\centering $Pressure (kPa)$} &
    \multicolumn{1}{|c|}{\centering $0$} &
    \multicolumn{1}{|c|}{\centering $30$} &
    \multicolumn{1}{|c|}{\centering $60$} &
    \multicolumn{1}{|c|}{\centering $90$} \\
    \hline   
    \multirow{3}{2.2cm}{$k_{lateral} (kPa/rad)$}
    & Plastic swab  & 1.107 & 1.115 & 1.117 & 1.119 \\
    & Wood swab  & 5.860 & 6.092 & 6.167 & 6.223 \\
    & Metal swab & 31.532 & 39.639 & 43.049 & 45.930 \\
    \hline
    \multirow{3}{2.2cm}{$k_{axial} (kPa/mm)$}
    & Plastic swab & 0.972 & 1.101 & 1.161 & 1.193 \\
    & Wood swab & 2.856 & 4.364 & 5.4659 & 6.269 \\
    & Metal swab & 3.010 & 4.735 & 6.061 & 7.064 \\
    \hline
\end{tabular}
\end{center}
\end{table}

\begin{figure}[!tb] 
	\vspace{-0.0cm}   
	\setlength{\abovecaptionskip}{-0.0cm}    
	\setlength{\belowcaptionskip}{-10cm}    
    \centering
    \includegraphics[width=\linewidth]{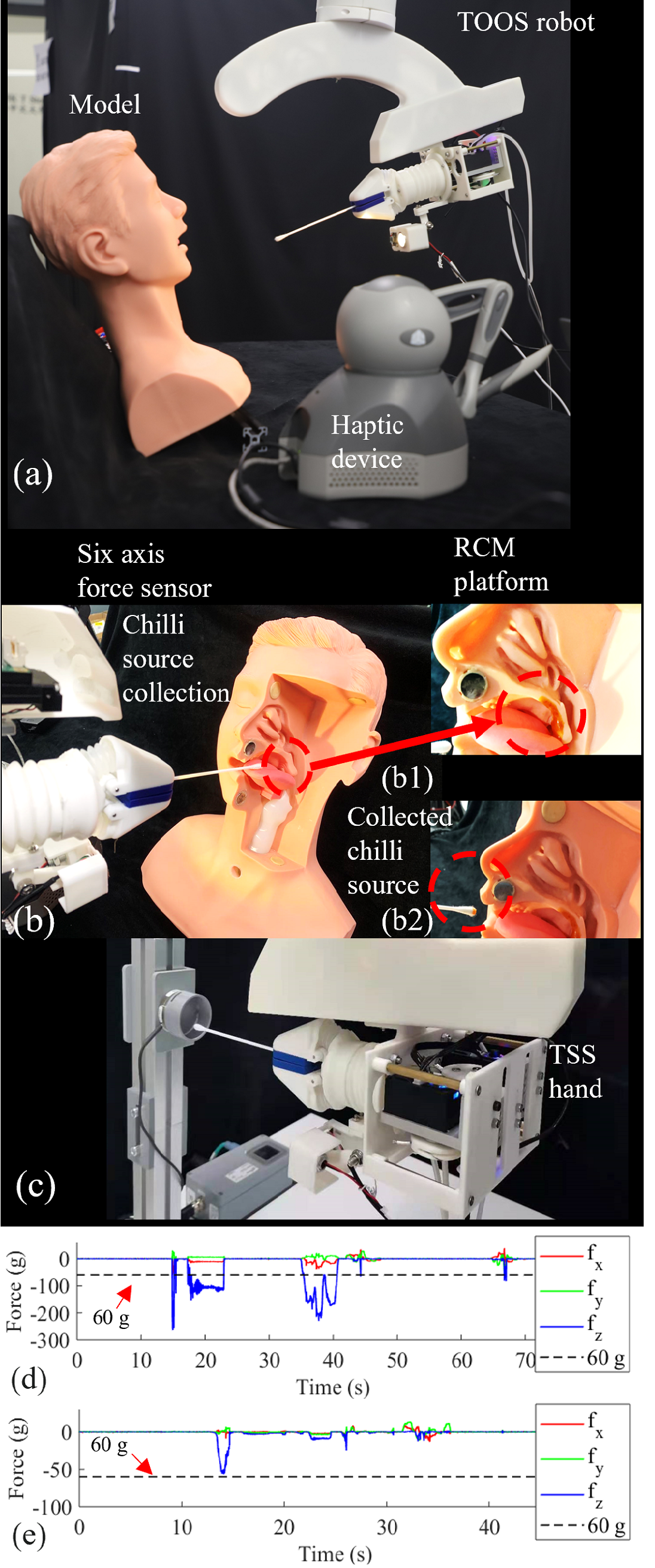}
    \caption{
        OP sampling success rate validation on nasopharyngology model and real-time sampling force measurement. 
        (a) System setup for sampling success rate test. 
        (b,b1,b2) Chilli sauce smeared on throat for effective sampling contact evaluation. 
        (c) Real-time sampling force test setup.
        (d) Real-time sampling force when the VF constraint is disabled.
        (e) Real-time sampling force when the VF constraint is enabled.
    }
    \label{fig:f13}
	\vspace{-0.2cm}   
\end{figure}

\begin{figure*}[!tb] 
	\vspace{-0.0cm}   
	\setlength{\abovecaptionskip}{-0.0cm}    
	\setlength{\belowcaptionskip}{-10cm}    
    \centering
    \includegraphics[width=\linewidth]{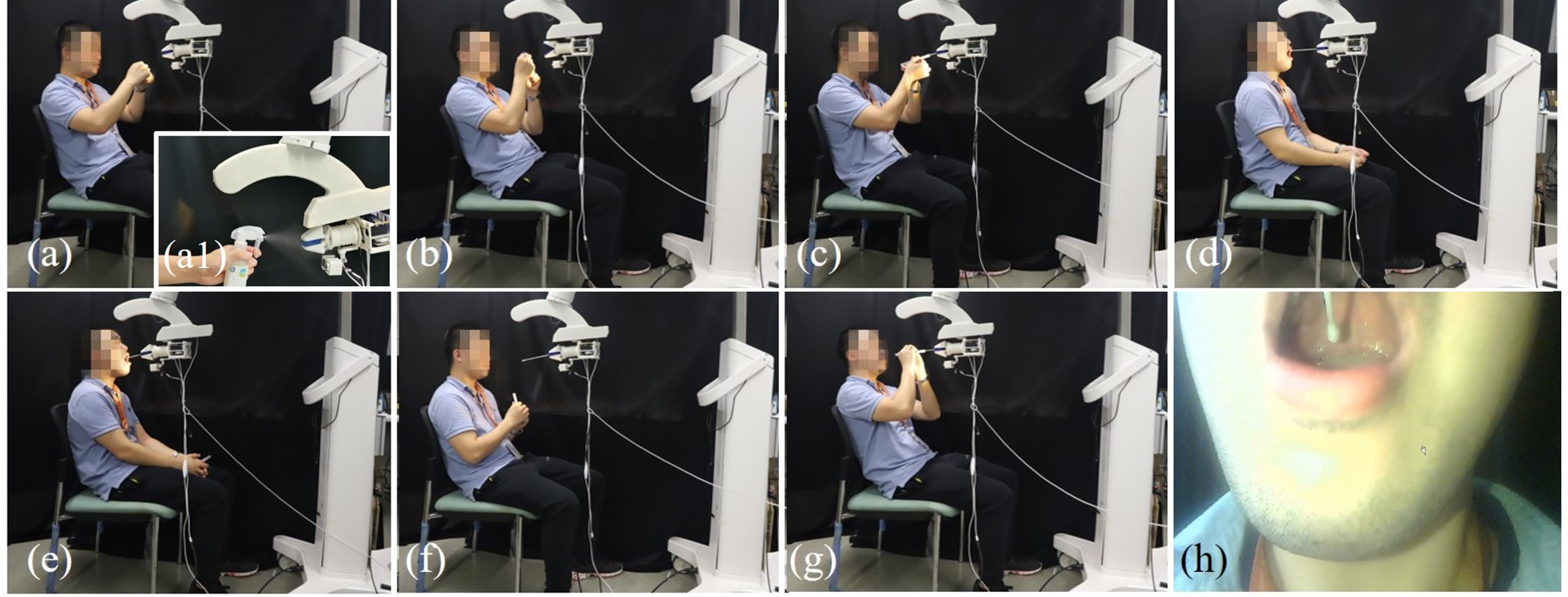}
    \caption{
        Human user OP sampling demonstration. 
        (a-a1) Sampling preparation and TSS hand can be directly sprayed by alcohol due to its waterproof feature.
        (b-c) Human-assisted swab replacing. 
        (d) Swab insertion into the human mouth. 
        (e) Swab teleoperation for specimen collection. 
        (f) Swab withdrawal after specimen collection. 
        (g) Human assisted swab collection. 
        (h) Terminal camera feedback when swab inserting into the mouth. 
    }
    \label{fig:f14}
	\vspace{-0.2cm}   
\end{figure*}

\subsection{TSS Hand Variable Stiffness Validation}

The second experiment were performed to determine the variable stiffness model of the TSS hand and to validate the force exertion safety of robotic sampling. 
Firstly, we measured the change in the lateral and axial forces of the TSS hand, specifically the soft wrist, with respect to the deflection at different pneumatic inputs to obtain the empirical relationship between the stiffness and the pneumatic input. 
Then, we measured the stiffness of the swabs individually. 
Three kinds swabs with different stiffness in plastic, wood, and iron material were used in this experiment.
Finally, the effective sampling stiffness was measured and compared with the modeling result. 
Figures \ref{fig:f11} (a) and \ref{fig:f11} (b) show the experimental setup for the TSS hand stiffness measurement. 
Both the axial and lateral stiffness were measured under different input pressures of 0, 30, 60, 90 kPa respectively. 
Results of the relationship between stiffness and pressure are depicted in Figs. \ref{fig:f12} (a) and \ref{fig:f12} (b), and the fitted relationship functions as shown in Figs. \ref{fig:f12} (c) and \ref{fig:f12} (d) can be used to adjust the stiffness of the robot as discussed in Section IV(D). 
The stiffness of an individual swab (without robot) was also tested at its terminal point. 
For example, the terminal interaction force of plastic swab in lateral and axial directions are depicted in Fig 12(e-f) with the comparison to the model value.
The plastic, wood, and iron swabs' axial stiffness were determined to be around 1.2851, 13.2447, and 17.5054 N/mm respectively while their lateral stiffness was around 1.1287, 6.5318, 70.6099 N/rad. 
Results shown in Table. II demonstrate that the wrist stiffness modulated by input pressure can effectively adjust the terminal interaction stiffness when mounting different swabs. 
The result also shows that the larger the terminal tool's stiffness, the larger the variable stiffness range.

Furthermore, in a real sampling process, testee sometimes will cough and there is a human-robot collision potential. The adaptable stiffness of TSS hand and its inherent compliance also provide human-robot collision safety protection.

\subsection{OP Sampling Success Rate Experiment}

To evaluate the sampling success rate, a sampling test on an oropharyngeal phantom model was performed using the TOOS robot and a plastic swab. 
The system setup is illustrated in Fig. \ref{fig:f13} (a) where an oropharyngeal phantom model was put on the table in front of the TOOS robot. 
The sampling success criterion is the effective contact between the swab and the surface of the oropharyngeal, which can be confirmed by the sample transfer from the oropharyngeal to the swab tip. 
The reddish yellow chilli sauce was used as the sampling target representing the potential virus attached to the surface of human OP tissue, as depicted in Fig. \ref{fig:f13} (b-b1).
The chilli source was smeared on the throat surface 

For the stiffness pre-adjustment, the stiffness was pre-adjusted to the maximum level with 90 kPa input pressure.

Firstly, the RCM platform with the swab attached to the TSS hand was placed near the mouth of the phantom model. 
Then, the RCM $J_1$ (insertion DOF) was actuated to insert the swab into the oral cavity. 
After the terminal of the swab reached the intended target, the RMC $J_1$ was locked by a limiting stopper. 
The TSS hand was then actuated via the teleoperation master device by the operator. 
The swab was maneuvered to be in contact with the throat surface. 
After the swab maneuvering, the TSS hand returned to its original home configuration state and locked by the limiting stopper. 
RCM $J_1$ was then unlocked and actuated to withdraw the swab from inside the mouth. 
A successful sampling cycle was evaluated by whether the swab surface contains the chilli source, as depicted in Fig. \ref{fig:f13} (b2). 
We performed the sampling tests for 20 times and collected the chilli source on all the 20 swabs successfully. 

The results prove that the high sampling success rate was ensured by the vision feedback-based teleoperation with a dexterous TSS hand.

As for the durability test, we have tested the motion accuracy and grasping robustness after 50 rounds of random sampling cycles. 
The prototype TSS hand remained dexterity in wrist motion and grasping robustness in swab picking without obvious difference from its original state. 
We have repeated this test two months after the orginal assembly of TSS hand, and TSS hand is still effective and provides similar performance to its original state, which validates the durability of the prototype TSS hand. 
Further durability enhancement can be achieved by material refinement for product level TSS hand fabrication.

\subsection{Tele-operation Sampling Force Experiment}

To test the real-time sampling force and the effectiveness of VF constraint, a sampling force experiment was processed as shown in Fig 13(c). 
A six axis force sensor, ATI Nano43 (ATI Industrial Automation), was used to record the real-time sampling force in X, Y, and Z directions. Tele-operated sampling was processed on an oral cavity model at 40 mm diameter, which was connected on the force sensor. 
When the VF constraint was disabled, the sampling force was easy to exceed the comfortable sampling threshold, 60 g sampling force, as shown in Fig 13(d). 
When the VF constraint was enabled when the swab was inserted into the oral cavity model, the sampling force can be effectively restricted in the comfortable zone, 0-60 g. 
The data of the experimental results proved the efficacy of the VF constraint of our sampling system.  

\subsection{Human User Demonstration} 

After the sampling safety and effectiveness validation, we presented real human user sampling to demonstrate the specific steps of the sampling process. 
The system setup is depicted in Fig. \ref{fig:f14} (a) where a volunteer testee was seated on a desk in front of the TOOS robot. 
The whole sampling procedure is illustrated in Figs. \ref{fig:f14} (a)-\ref{fig:f14} (h). 
The specific procedure is provided below.

\textbf{TOOS OP sampling procedure:}

\textbf{Step\_1 Sampling preparation and TSS hand disinfection:}
The testee takes a seat and the operator turns on the TOOS robot as shown in Fig. \ref{fig:f14} (a). 
The stiffness of the robot was set at the highest level with 90 kPa pneumatic input to ensure teleoperation accuracy. The TSS hand can be disinfected as shown in Fig14 (a1).

\textbf{Step\_2 Testee-assisted swab mounting:}
Testee provides the swab to the TSS hand and the hand grasps the swab 
firmly as shown in Figs. \ref{fig:f14} (b) and \ref{fig:f14} (c). 

\textbf{Step\_3 Locking the TSS hand:}
The testee opens the mouth 
in front of the TSS hand waiting for sampling. The TSS 
hand is kept in the original straight configuration. 

\textbf{Step\_4 Swab insertion and VF range selection:}
$J_1$ of the RCM platform is actuated to insert the swab into the testee's mouth, as shown in Fig. \ref{fig:f14} (d). During this process, the operator can adjust the VF range in operation UI, Fig 6(c), based on the camera feedback of oral configuration as shown in Fig 14(h).

\textbf{Step\_5 Tele-operated sampling:}
The operator stops the insertion when the swab approaches the target based on visual feedback. 
Then the TSS hand is unlocked. 
Tele-operation is applied to collect the sample from the OP tissue, as shown in Fig. \ref{fig:f14} (e).

\textbf{Step\_6 Locking the TSS hand:}
After the collection is achieved based on visual feedback, the TSS hand is locked again and returns to its original straight configuration.

\textbf{Step\_7 Swab exertion:} 
$J_1$ of RCM platform is then actuated to withdraw the swab from the human mouth, as shown in Fig. \ref{fig:f14} (f).

\textbf{Step\_8 Testee-assisted swab collection:}
After the swab is withdrawn, the testee is required to collect the swab, as shown in Fig. \ref{fig:f14} (g). 
After that, a complete round of OP sampling is achieved.

The whole sampling process was performed without report from the human user of any uncomfortable contact generated by the robot. 
There were three operators involved in the tele-operation sampling, with one well-trained Doctor and two robotic researchers. The Doctor who had jointed real COVID-19 OP swab sampling before the tele-operation test and had experience in tele-operated medical device. The Doctor equips well-trained knowledge on the OP swab sampling and experience on the manual sampling strategy. Two robotics researchers have no real OP sampling experience, and one of them has robot tele-operation experience. After a rapid overhand operation, all the three operators can understand and grasp the operation method of the tele-operation system. The swab can be tele-operated at will by the operators. The VF constraint was also acceptable for all three operators to restrict their over large motion. 

The overall sampling time is around 30 seconds for our OP sampling robot. 
The human sampling time is around 5 to 10 seconds. 
Voice communication time is around 10 seconds for testee to fit the position and prepare for sampling. 
In terms of the swab manipulation, swab collecting the specimen after insertion, the consumed time are quite similar between manual and robot operation. The major difference is at the insertion and extraction speed by the RCM platform, in which the maximum speed is restricted. 
This speed can be enhanced by replacing a faster insertion and extraction mechanism.
The conversation between medical worker and testee can be achieved by the integrated microphone and audio devices. 
The swab picking, releasing, and collection can be achieved with simple human assistance. 

\section{CONCLUSION AND FUTURE WORK}
In this paper, we present soft robotic technology is a viable solution for safe and effective robotic OP swab sampling. 
Based on human sampling observation, we concluded three physical interaction components (arm, wrist, and hand) and two sensing components (visual hand haptic feedback) were mainly involved in the manual OP sampling. 
The arm motion is easy to be replaced by robot arm or multi-DOF robotic mechanism. 
Thus, we developed a 3-DOFs RCM platform to replace the human arm to provide 
global positioning of the terminal sampling equipment. 
However, building a robotic hand for dexterous swab manipulation in a confine space is challenging. 
To tackle this challenge, a novel TSS hand consisting of an universal soft wrist and a soft gripper was proposed to replace the human wrist and human hand. 
TSS hand is one of the most dexterous soft hand dedicated for disposable swab manipulation and provides human hand comparable dexterity and workspace. 
The stiffness of the soft wrist can be further pre-adjusted by pneumatic input to ensure both safe and forceful interaction. 
We applied a teleoperation controller with hybrid virtual fixture constrain for the sampling control. 
Comprehensive experiments were performed to validate the function of the sampling system, including accuracy of the teleoperation framework, variable stiffness of the TSS hand, and operation safety of robotic sampling.
The soft robotics-enabled OP sampling system using existing disposable swab demonstrates promising potential to automate OP sampling process with excellent safety guarantee for both medical staffs and testees.

In the future work, we will be seeking ethics approval to perform in-vivo test using our OP sampling system, further refining the TSS hand, integrating sensors to detect contact of the specimen, detect the oral configuration in real-time, refining the hybrid virtual fixture controller, and transferring the TSS hand to two collaborative arms for dual-arm manipulation. 
In terms of disinfection, disposable shielding cover and UVC sterilization will be considered.
Furthermore, the teleoperation controller can be improved by integrating computer vision assistance and artificial intelligence  enabled by human behaviour learning.




\bibliographystyle{./IEEEtran}
\bibliography{./ref}

\end{document}